\documentclass[11pt]{article}

\usepackage[preprint]{acl}

\usepackage{times}
\usepackage{latexsym}
\usepackage{booktabs}
\usepackage[T1]{fontenc}
\usepackage[utf8]{inputenc}

\usepackage{microtype}

\usepackage{inconsolata}
\usepackage{listings}

\lstdefinestyle{promptstyle}{
    basicstyle=\ttfamily\scriptsize,
    breaklines=true,
    breakatwhitespace=true,
    columns=fullflexible,
    keepspaces=true,
    showstringspaces=false,
    frame=single,
    linewidth=\columnwidth,
}

\usepackage{graphicx}

\newcommand{\ie}{i.e.\ }

\title{GAVEL: Grounded Caption Error Verification and Localization}

\author{
Zixian Gao \quad Atsushi Hashimoto \quad Kuniaki Saito \\
OMRON SINIC X Corporation \\
\texttt{\{zixian.gao, atsushi.hashimoto, kuniaki.saito\}@sinicx.com}
}

\begin{document}
\maketitle

\begin{abstract}
% Vision–language models (VLMs) often produce hallucinated or inconsistent outputs, where text and images are not properly aligned. Addressing this issue requires not only detecting misalignment but also explaining the discrepancy and localizing its visual evidence. We introduce GAVEL, a task that jointly performs verification, explanation, and localization for image–text pairs, and provide a corresponding dataset and benchmark to enable systematic evaluation. We propose \methodname, a unified framework that leverages hybrid supervision—combining synthetic and human-annotated data—to jointly solve all three tasks. Our approach transforms open-vocabulary detection datasets into GAVEL-style supervision, enabling scalable training without exhaustive annotation. Experiments show that even strong closed-source models struggle on GAVEL, while our method significantly outperforms existing approaches. Notably, hybrid supervision yields consistent gains across all tasks.
Vision–language models (VLMs) often produce hallucinated or inconsistent outputs, where text and images are not properly aligned. Addressing this issue requires not only detecting misalignment but also explaining the discrepancy and localizing its visual evidence. We introduce \textbf{GAVEL} (Grounded Caption Error Verification and Localization), a task that jointly addresses verification, explanation, and localization for image–text pairs. To support systematic evaluation, we also present a corresponding dataset and benchmark. We further train a supervised baseline on the human-annotated training split to assess whether GAVEL provides learnable supervision for these abilities. Experiments show that even strong closed-source models struggle on GAVEL, while the supervised baseline yields consistent improvements across grounding and explanation metrics. We will publish the dataset and the evaluation code upon acceptance.
\end{abstract}

\section{Introduction}
Recent vision–language models (VLMs) can generate detailed textual descriptions from images~\cite{li2023blip,chen2024sharegpt4v,llama4,wang2024cogvlm,bai2025qwen3} and synthesize images from textual prompts~\cite{sdlarge,deng2025emerging}. Despite their impressive generative capabilities, these models frequently produce hallucinated content: captions may mention objects or attributes absent from the image, while generated images may fail to faithfully reflect the input text. Such inconsistencies raise concerns about reliability, interpretability, and downstream usability in real-world applications.

Understanding inconsistencies between images and text is a fundamental problem underlying many vision–language tasks. Image classification, object detection, segmentation, and image–text retrieval can all be viewed as measuring consistency between visual evidence and textual descriptions~\cite{radford2021learning,li2022grounded,liang2023open}. Similarly, in human–robot interaction and instruction following, systems must identify whether textual instructions correctly correspond to observed visual environments~\cite{shridhar2020alfred,kwok2026scaling}. While prior studies have explored hallucination detection~\cite{saito2026haldec,gunjal2024detecting,wada2025zina} or factuality evaluation, existing benchmarks primarily focus on coarse correctness judgments or text-only correction, providing limited insight into why a mismatch occurs and where it is grounded in the image.
\begin{figure}[t]
\centering
\includegraphics[width=0.46\textwidth]{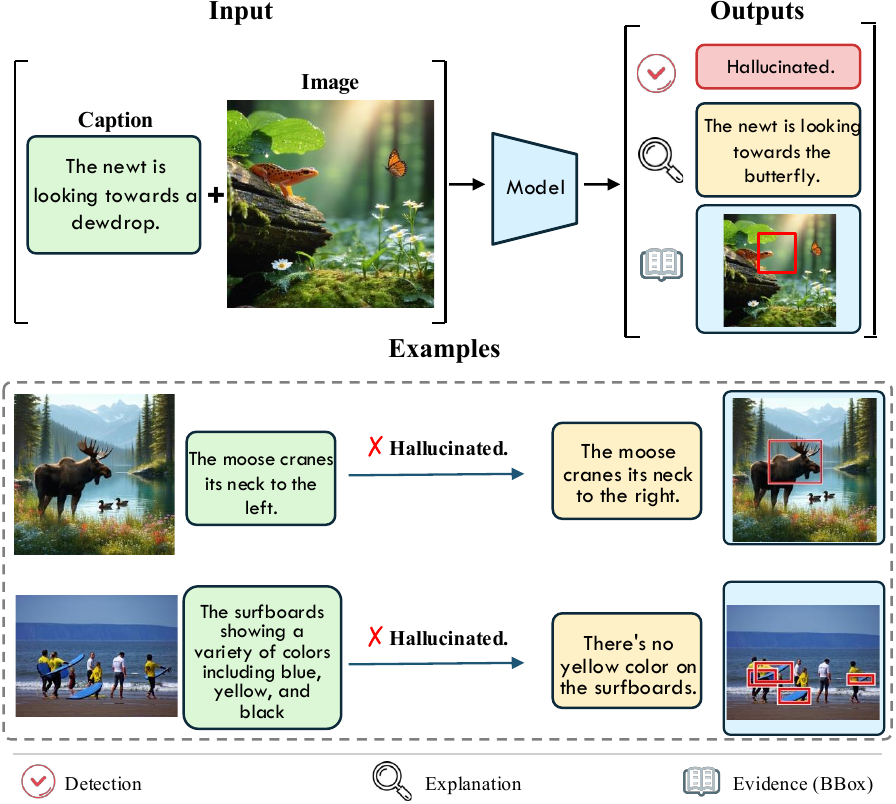}
\caption{Overview of the proposed GAVEL task. Given an image–text pair containing an inconsistency, the goal is to explain the discrepancy in natural language and localize the corresponding visual evidence. The visual evidence region is highlighted in \textcolor{red}{red}.}
\label{fig:teaser}
\end{figure}

To address this limitation, we introduce GAVEL (Grounded cAption Error Verification and Localization), a new benchmark for grounded vision–language inconsistency understanding. Given an inconsistent image–text pair, the task requires models to: (1) explain the inconsistency in natural language, and (2) localize the visual regions related to the discrepancy. By jointly evaluating explanation and localization, GAVEL enables a more fine-grained and interpretable analysis of vision–language alignment than prior benchmarks.
The proposed benchmark offers several advantages. First, it evaluates whether a model can identify the underlying cause of a mismatch rather than merely predicting that an inconsistency exists. Second, grounding explanations in visual regions provides interpretable evidence that facilitates human understanding and diagnosis. Third, the task naturally supports practical applications such as automatic caption correction, hallucination analysis, and iterative refinement in generative systems, where identifying specific erroneous components is more useful than coarse binary feedback.
To support this benchmark, we curate a dataset with diverse image–text inconsistencies collected from both image-to-text and text-to-image generation pipelines. The dataset contains grounded annotations describing the inconsistency together with corresponding visual evidence, enabling systematic evaluation of grounded vision–language understanding.

In addition, we establish GAVEL, including benchmark protocols and evaluation metrics tailored to grounded inconsistency explanation and localization. The benchmark jointly measures explanation quality and localization performance, enabling detailed analysis of model capabilities in identifying and grounding vision–language discrepancies. We believe GAVEL will serve as a valuable testbed for developing more reliable, interpretable, and grounded vision–language systems.
Our contributions are summarized as follows:
\begin{itemize}
\item We introduce GAVEL, a new benchmark for grounded vision–language inconsistency understanding that jointly requires discrepancy explanation and visual localization, providing a foundation for future research on reliable and interpretable vision–language alignment.
\item We construct a dataset containing diverse image–text inconsistencies from both image-to-text and text-to-image generation pipelines, together with grounded annotations and explanatory descriptions.
\end{itemize}

\section{GAVEL Construction}

% Our goal is constructing a dataset with four elements, where they are paired together and form a single data instance:
% \((I, D, E, G)\), where \(I\) denotes the image, \(D\) the single-sentence textual description, \(E\) the textual explanation of any misalignment, and \(G\) the visual grounding information, \ie, bounding box coordinates. In this section, we explain the detailed process to construct our dataset and its stats. 
Our goal is to construct a dataset in which each instance is represented as
\((I, D, E, G)\), where \(I\) denotes an image, \(D\) denotes a single-sentence textual description, \(E\) denotes a textual explanation describing any image-text misalignment, and \(G\) denotes the visual grounding information, \ie, bounding-box coordinates. In this section, we describe the dataset construction process and present its statistics.
% \begin{figure}[t]
%     \centering
%    \includegraphics[width=0.5\textwidth]{images/fig_ann.pdf}
%     \caption{Overview of the pipeline to construct dataset.}
%     \label{fig:dataset_construction}
% \end{figure}

\subsection{Data Preparation}
The challenge in constructing the dataset is how to collect hallucinated image-caption pairs. For the test split, we employ the negative pairs from Haldec~\cite{saito2026haldec}, which provide sentence-level hallucination existence annotation, wherein we employ eight models, including Qwen2~\cite{wang2024qwen2}, Llama-4~\cite{llama4}, LLaVA-Next~\cite{li2024llavanext-strong}, ShareGPT~\cite{chen2024sharegpt4v}, CogVLM~\cite{wang2024cogvlm}, GPT-4o~\cite{chatgpt_citation}, Stable-Diffusion~\cite{sdlarge}, and an image generation model used in GPT.

To construct the training set, we design a semi-automatic pipeline that leverages VLMs to generate and select likely negative image–caption pairs, which are then sent to human annotators for further annotation. An overview of the data construction process is provided in the appendix. We first generate captions using a vision–language model (VLM) or a text-to-image model, and then send the resulting image–caption pairs to GPT-5 mini~\cite{chatgpt_citation} to evaluate their alignment.
The intuition is that pairs with lower alignment scores are more likely to be negative examples. We subsequently forward these low-scoring pairs to human annotators to obtain complete annotations.

\subsection{Annotation}
%\noindent\textbf{Details of the annotation.} 
Given image–sentence pairs, we ask annotators to (i) explain any textual misalignment when the pair is incorrect, and (ii) provide a bounding box corresponding to the image region that grounds the error. To ensure annotation quality, we employ a three-stage verification process. First, each sentence is assigned to both an annotator and a reviewer, with any errors corrected during review. Subsequently, additional reviewers randomly inspect samples to further ensure the overall quality of the annotations.

\subsection{Stats of the dataset}
\noindent\textbf{Overall Stats.}
Our human-annotated training dataset consists of 30,309 images with 35,249 sentences with detailed annotations in total, while the testing dataset consists of 2,569 images with 5,606 sentences. Thus, the training is conducted on 35,000 pairs, and testing is conducted on 5,000 pairs. 
Also, we confirm that our way of the negative pairs for the training data is working well, considering that the ratio of "No hallucination" pairs is only 7.3\%. 

\noindent\textbf{Sentence Stats.}
Each sentence consists of 24.82 words on average, and the vocabulary covers 27,173 unique words.

\begin{table*}[t]
\centering
\caption{Performance of existing models on GAVEL, including visual grounding accuracy and LLM-based scores for textual explanations across different hallucination categories. Textual explanation scores are scaled to 0--100.}
\label{tab:results}
\resizebox{\textwidth}{!}{%
\begin{tabular}{lcccccccccccccc}
\toprule
\textbf{Method} 
& \multicolumn{3}{c}{\textbf{Visual Grounding}} 
& \multicolumn{10}{c}{\textbf{Textual Explanation (LLM Score)}} \\
\cmidrule(lr){2-4} \cmidrule(lr){5-15}
& \textbf{A@0.3} & \textbf{A@0.5} & \textbf{A@0.7} 
& \textbf{Avg} 
& \textbf{Act} 
& \textbf{Attr} 
& \textbf{Color} 
& \textbf{Loc} 
& \textbf{Obj} 
& \textbf{Qty} 
& \textbf{State} 
& \textbf{Size} 
& \textbf{No hall.}
& \textbf{Other} \\
\midrule

GPT-5             
& \textbf{34.1} & 17.4 & 6.5  
& \textbf{70} & \textbf{72} & 72 & 66 & \textbf{68} & \textbf{78} & 72 & \textbf{76} & \textbf{70} & 58 & 68 \\

GPT-5-mini        
& 32.5 & 14.7 & 4.2  
& \textbf{70} & 70 & \textbf{72} & \textbf{72} & 68 & 76 & \textbf{72} & 68 & 68 & 68 & \textbf{78} \\

InternVL-3.5VL-8B  
& 17.3 & 10.3 & 6.0  
& 44 & 38 & 40 & 44 & 38 & 54 & 36 & 36 & 30 & \textbf{82} & 42 \\

InternVL-3.5VL-30B 
& 23.5 & 15.8 & 9.4  
& 52 & 48 & 48 & 54 & 48 & 60 & 48 & 50 & 40 & 76 & 52 \\

Qwen3-VL-8B       
& 15.1 & 12.1 & 9.1  
& 54 & 58 & 50 & 52 & 52 & 60 & 46 & 52 & 42 & 74 & 46 \\

Qwen3-VL-30B      
& 30.8 & \textbf{22.8} & \textbf{16.4} 
& 62 & 66 & 60 & 64 & 60 & 70 & 52 & 66 & 58 & 64 & 52 \\

\bottomrule
\end{tabular}%
}
\end{table*}

\begin{figure}[h]
    \centering
   \includegraphics[width=0.46\textwidth]{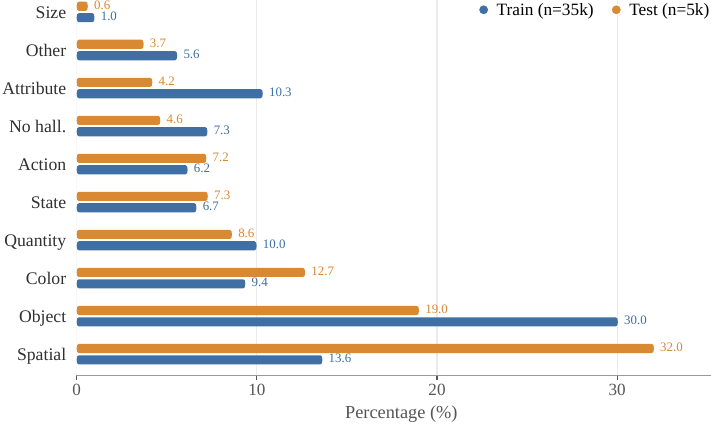}
    \caption{Overview of hallucination type distribution on the test and train set, where the dark bars represent the training set and the light bars represent the test set.}
    \label{fig:hall_cate}
\end{figure}

\noindent\textbf{Bounding box stats.}
On average, 1.32 bounding boxes are assigned per image. This suggests that visual grounding is often concentrated on a single region, whereas annotators frequently provide annotations for multiple regions—for example, in cases of hallucinations during object counting. We further visualize the distribution of bounding box sizes in the appendix. The results confirm that the annotated boxes span both small and large regions, indicating that our task effectively evaluates a model’s localization ability across a wide range of spatial scales.

\noindent\textbf{Hallucination Categories.}
Figure~\ref{fig:hall_cate} illustrates the hallucination type distributions in the train and test sets. 
To obtain this categorization, we use GPT-5 to classify each hallucination instance into predefined hallucination types~\cite{saito2026haldec} based on the annotated caption error and its corresponding explanation. 
The resulting distribution shows that our test set covers diverse fine-grained hallucination types, making it suitable for robustly evaluating models' abilities in caption error verification, explanation, and localization.

\noindent\textbf{Examples.}
In Fig. ~\ref{fig:teaser}, we visualize examples of images, hallucinated captions, misalignment description, and visual grounding. More examples are provided in the appendix.

% \begin{figure}[t]
%     \centering
%    \includegraphics[width=0.5\textwidth]{images/train_test_type_ratio.pdf}
%     \caption{Overview of hallucination type distribution on the test and train set, where the dark bars represent the training set and the light bars represent the test set.}
%     \label{fig:dataset_construction}
% \end{figure}

% \input{latex/chapters/method}
\section{Experiment}

% 1. Our approach achieves better performance than existing methods on the main task.(hallucination grounding, hallucination detection)

% a. We need to create two separate tables to present our model’s performance on these two tasks.

% \noindent
% 2. Our two different training datasets indeed play different roles in the model’s performance.

% a. Use a line chart to illustrate the impact of different training data on model performance

% \noindent
% 3. To demonstrate the effect of balancing the number of data samples across different hallucination types in our dataset.

% a. Use a bar chart to illustrate the impact of different fusion layers on model performance.

% \noindent
% 4. Validation of the effectiveness of different modules in the proposed method and sensitivity to hyperparameters.(1). Effect of different bin numbers. (2). Effect of different MultiFusion fusion strategies.

% \noindent
% 5. Show some visualization results to show our model's performance directly.

% a. draw some figures.

\subsection{Experimental Settings}

\noindent
We evaluate our method against a diverse set of strong baselines, including both proprietary and open-source vision-language models across different model scales.

\noindent
\textbf{Closed-source Models.}
We consider recent state-of-the-art proprietary models, including GPT-5 and GPT-5-mini ~\cite{singh2025openai}.

\noindent
\textbf{Open-source Models.}
We include a diverse set of open-source vision-language models across different parameter scales. For comparable-scale evaluation, we consider two representative 8B models: InternVL3.5-8B-Instruct~\cite{wang2025internvl3} and Qwen3-VL-8B-Instruct~\cite{bai2025qwen3}. To further examine scaling effects, we include larger models such as InternVL3.5-30B-A3B-Instruct and Qwen3-VL-30B-A3B-Instruct. 
% \noindent
% \textbf{Evaluation Tasks.}
% We conduct experiments on two complementary hallucination-related tasks. The first is our proposed \textit{Hallucination Grounding} task, which requires models to (1) identify hallucinated content in captions and (2) localize corresponding visual evidence in the image. The second is HalDec-Bench~\cite{saito2026haldec}, which focuses on binary hallucination detection in captions. Together, these tasks evaluate both hallucination detection and explanation capabilities. All models are evaluated under consistent prompting and decoding settings to ensure fair comparison.

\noindent
\textbf{Evaluation Metrics.}
We evaluate bounding box predictions using IoU-based accuracy under thresholds of 0.3, 0.5, and 0.7.  For textual explanations, we use GPT-5 as an automatic evaluator to assign scores from 0 to 5, which are multiplied by 20 and reported on a 0--100 scale for both average and category-wise performance.

\begin{table}[h]
\centering
\caption{Effectiveness of model training on human-annotated data.}
\label{tab:epoch}
\small
\setlength{\tabcolsep}{5pt}
\renewcommand{\arraystretch}{1.05}
\begin{tabular}{ccccc}
\toprule
\textbf{Model} & \textbf{A@0.3} & \textbf{A@0.5} & \textbf{A@0.7} & \textbf{LLM Avg} \\
\midrule
Base Model & 15.1 & 12.1 & 9.1 & 54 \\
% 5 & 30.4 & 19.6 & 11.0 & 2.8 \\
% 6 & 35.5 & 23.6 & 14.1 & 2.8 \\
Ours & \textbf{42.6} & \textbf{30.2} & \textbf{18.4} & \textbf{58} \\
\bottomrule
\end{tabular}
\end{table}

\subsection{Results}
\textbf{Hallucination Grounding Results. }
To evaluate current multimodal models on the proposed Hallucination Grounding benchmark, we compare representative closed- and open-source models in Table~\ref{tab:results}. The benchmark measures visual grounding with A@0.3, A@0.5, and A@0.7, and textual explanation quality with LLM scores across hallucination categories. Results show that Hallucination Grounding remains challenging. GPT-5 achieves the best A@0.3, indicating stronger coarse localization, while Qwen3-VL-30B performs best at stricter IoU thresholds, with the highest A@0.5 and A@0.7. For textual explanations, GPT-5 obtains the highest LLM score, followed by Qwen3-VL-30B. Category-level results reveal notable variation across hallucination types. Overall, our benchmark offers a diagnostic setting for analyzing hallucination localization and explanation quality.
% To evaluate current multimodal models on the proposed Hallucination Grounding benchmark, we compare closed-source and open-source models, as shown in Table~\ref{tab:results}. The benchmark evaluates two complementary abilities: visual grounding, measured by A@0.3, A@0.5, and A@0.7, and textual explanation quality, measured by LLM scores across hallucination categories.
% The results show that Hallucination Grounding remains challenging for existing models. GPT-5 achieves the best A@0.3 accuracy, suggesting stronger coarse localization ability, while Qwen3-VL-30B performs best under stricter IoU thresholds, achieving the highest A@0.5 and A@0.7 scores. For textual explanations, GPT-5 obtains the highest average LLM score, followed by Qwen3-VL-30B. The category-level breakdown reveals that model performance varies notably across different hallucination types.
% Overall, these results demonstrate that our benchmark provides a challenging and diagnostic evaluation setting. It enables fine-grained analysis of both where hallucinations occur in the image and how well models can explain them.

\noindent
\textbf{Analysis of Training Set Learnability.}
We further examine whether the training split can support model learning. 
Using the VisionLLM framework~\cite{wang2023visionllm} as the base model, we fine-tune it on our training set and evaluate the trained model on the test set. As shown in Table~\ref{tab:epoch}, we report the epoch-7 checkpoint.
Compared with the base model without training, our trained model achieves consistent improvements across grounding metrics. 
This shows that the dataset contains learnable and generalizable grounding signals, suggesting that our benchmark provides both reliable evaluation data and effective training supervision for future model development.

\begin{figure}[h]
    \centering
   \includegraphics[width=0.48\textwidth]{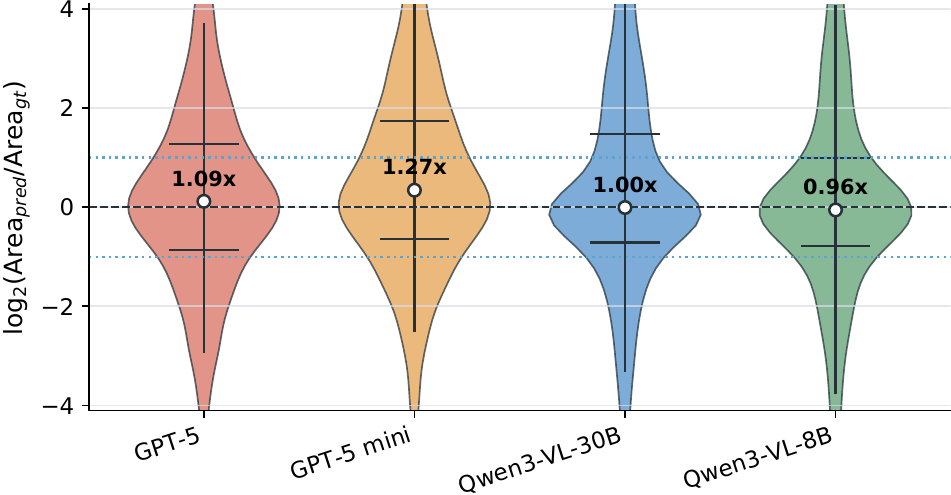}
    \caption{Distribution of predicted-to-ground-truth bounding-box area ratios. The y-axis shows $\log_2(\mathrm{Area}(B_{pred})/\mathrm{Area}(B_{gt}))$: 0 indicates equal area, 1 twice the ground-truth area, and -1 half the ground-truth area.}
    \label{fig:grounding_box_relation}
\end{figure}

\vspace{-5mm}

\noindent
\textbf{Failure Analysis of Visual Grounding.}
We further analyze whether visual grounding errors are related to bounding-box size bias.
For each valid prediction, we compute the predicted-to-ground-truth area ratio, $\mathrm{Area}(B_{pred})/\mathrm{Area}(B_{gt})$, for GPT-5, GPT-5 mini, Qwen3-VL-30B, and Qwen3-VL-8B.
Figure~\ref{fig:grounding_box_relation} shows the log-scale distribution of this ratio, where 0 indicates equal box size.
Most models have median ratios close to 1.00, suggesting no systematic over- or under-sizing.
However, GPT-5 mini tends to produce larger boxes, while Qwen3-VL-8B is more concentrated around the ground-truth size.
Together with the grounding accuracy results, this suggests that grounding failures stem not only from box-size bias, but also from inaccurate localization.
% \vspace{-5mm}

\begin{figure}[h]
\centering
\includegraphics[width=0.49\textwidth]{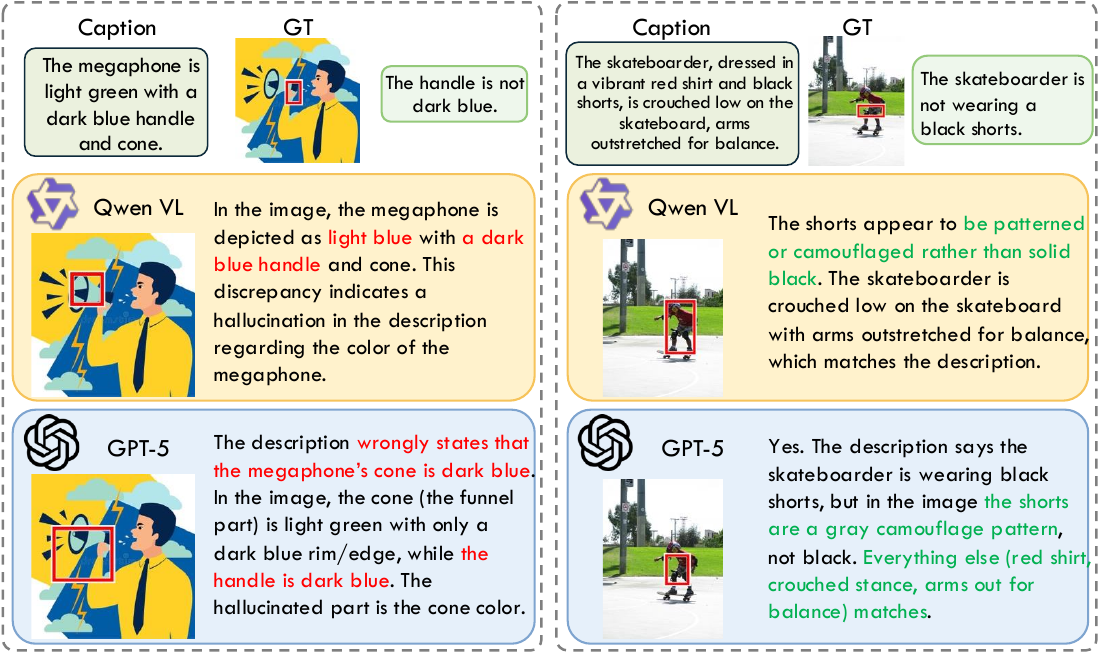}
\caption{GAVEL qualitative examples for hallucination detection and localization.}
\label{fig:vis}
\end{figure}
\vspace{-3mm}
% \noindent

\noindent
\textbf{Qualitative comparison.}
% Figure~\ref{fig:vis} presents qualitative comparisons between InternVL and GPT-5 on hallucination detection and grounding. For each example, we show the input caption, the ground-truth hallucination explanation and bounding box, and predictions from both models. The red boxes indicate localized visual evidence associated with the hallucinated content. The highlighted text marks the key claim identified by each model. These examples show that different models succeed or fail along different dimensions: a model can identify the hallucinated phrase but localize an imprecise region, or provide a partially incorrect explanation despite predicting a plausible bounding box.
Figure~\ref{fig:vis} presents qualitative comparisons between InternVL and GPT-5 on hallucination detection and grounding. For each example, we show the input caption, ground-truth explanation and bounding box, and predictions from both models. Red boxes indicate localized visual evidence, while highlighted text marks the key claim identified by each model. These examples show that models may fail in different ways: identifying the hallucinated phrase but localizing an imprecise region, or giving a partially incorrect explanation despite a plausible bounding box.

\section{Conclusion}

GAVEL shows the need to move beyond image- or response-level hallucination assessment by jointly evaluating detection, explanation, and localization. Beyond benchmarking, it provides training data for improving hallucination understanding and grounding across diverse hallucination categories. This direction is crucial for building more reliable, interpretable, and practical vision-language systems.

\section{Limitations}

This work primarily focuses on GAVEL as a fine-grained benchmark for evaluating hallucination detection, explanation, and localization in vision-language models. Although we provide a training split and conduct preliminary experiments showing that the dataset contains learnable grounding signals, we do not fully explore its potential for model training. More systematic studies on training strategies, data scaling, data mixture design, and model architectures are left for future work.

\section*{Acknowledgements}
This work was supported by JST PRESTO, Japan, Grant Number JPMJPR2523. This work was partly achieved through the use of SQUID at D3 Center, The University of Osaka. 

\bibliography{custom}

@String(CVPR  = {IEEE Conf. Comput. Vis. Pattern Recog.})

@String(ECCV  = {Eur. Conf. Comput. Vis.})

@String(ICML  = {Int. Conf. Mach. Learn.})

@String(AAAI  = {AAAI})

@String(CVPR  = {CVPR})

@String(ECCV  = {ECCV})

@String(NIPS = {NeurIPS})

@String(ICML  = {ICML})

@article{kwok2026scaling,
  title={Scaling Verification Can Be More Effective than Scaling Policy Learning for Vision-Language-Action Alignment},
  author={Kwok, Jacky and Zhang, Xilun and Xu, Mengdi and Liu, Yuejiang and Mirhoseini, Azalia and Finn, Chelsea and Pavone, Marco},
  journal={arXiv preprint arXiv:2602.12281},
  year={2026}
}

@inproceedings{shridhar2020alfred,
  title={Alfred: A benchmark for interpreting grounded instructions for everyday tasks},
  author={Shridhar, Mohit and Thomason, Jesse and Gordon, Daniel and Bisk, Yonatan and Han, Winson and Mottaghi, Roozbeh and Zettlemoyer, Luke and Fox, Dieter},
  booktitle=CVPR,
  pages={10740--10749},
  year={2020}
}

@article{deng2025emerging,
  title={Emerging properties in unified multimodal pretraining},
  author={Deng, Chaorui and Zhu, Deyao and Li, Kunchang and Gou, Chenhui and Li, Feng and Wang, Zeyu and Zhong, Shu and Yu, Weihao and Nie, Xiaonan and Song, Ziang and others},
  journal={arXiv preprint arXiv:2505.14683},
  year={2025}
}

@inproceedings{liang2023open,
  title={Open-vocabulary semantic segmentation with mask-adapted clip},
  author={Liang, Feng and Wu, Bichen and Dai, Xiaoliang and Li, Kunpeng and Zhao, Yinan and Zhang, Hang and Zhang, Peizhao and Vajda, Peter and Marculescu, Diana},
  booktitle=CVPR,
  pages={7061--7070},
  year={2023}
}

@inproceedings{li2022grounded,
  title={Grounded language-image pre-training},
  author={Li, Liunian Harold and Zhang, Pengchuan and Zhang, Haotian and Yang, Jianwei and Li, Chunyuan and Zhong, Yiwu and Wang, Lijuan and Yuan, Lu and Zhang, Lei and Hwang, Jenq-Neng and others},
  booktitle=CVPR,
  year={2022}
}

@article{wang2023visionllm,
  title={Visionllm: Large language model is also an open-ended decoder for vision-centric tasks},
  author={Wang, Wenhai and Chen, Zhe and Chen, Xiaokang and Wu, Jiannan and Zhu, Xizhou and Zeng, Gang and Luo, Ping and Lu, Tong and Zhou, Jie and Qiao, Yu and others},
  journal=NIPS,
  volume={36},
  pages={61501--61513},
  year={2023}
}

@misc{saito2026haldec,
      title={HalDec-Bench: Benchmarking Hallucination Detector in Image Captioning}, 
      author={Kuniaki Saito and Risa Shinoda and Shohei Tanaka and Tosho Hirasawa and Fumio Okura and Yoshitaka Ushiku},
      year={2026},
      eprint={2603.15253},
      archivePrefix={arXiv},
      primaryClass={cs.CV},
      url={https://arxiv.org/abs/2603.15253}, 
}

@article{singh2025openai,
  title={Openai gpt-5 system card},
  author={Singh, Aaditya and Fry, Adam and Perelman, Adam and Tart, Adam and Ganesh, Adi and El-Kishky, Ahmed and McLaughlin, Aidan and Low, Aiden and Ostrow, AJ and Ananthram, Akhila and others},
  journal={arXiv preprint arXiv:2601.03267},
  year={2025}
}

@article{bai2025qwen3,
  title={Qwen3-vl technical report},
  author={Bai, Shuai and Cai, Yuxuan and Chen, Ruizhe and Chen, Keqin and Chen, Xionghui and Cheng, Zesen and Deng, Lianghao and Ding, Wei and Gao, Chang and Ge, Chunjiang and others},
  journal={arXiv preprint arXiv:2511.21631},
  year={2025}
}

@article{wang2024cogvlm,
  title={Cogvlm: Visual expert for pretrained language models},
  author={Wang, Weihan and Lv, Qingsong and Yu, Wenmeng and Hong, Wenyi and Qi, Ji and Wang, Yan and Ji, Junhui and Yang, Zhuoyi and Zhao, Lei and XiXuan, Song and others},
  journal=NIPS,
  year={2024}
}

@misc{sdlarge, 
  title        = {Stable Diffusion 3.5 Large},
  author={{Stability AI}},
  howpublished = {https://stability.ai/news/introducing-stable-diffusion-3-5},
  year         = {2024}
}

@misc{llama4, 
  title        = {The Llama 4 herd: The beginning of a new era of natively multimodal AI innovation},
  author={Meta.AI},
  url = {https://ai.meta.com/blog/llama-4-multimodal-intelligence/},
  year         = {2025}
}

@misc{li2024llavanext-strong,
    title={LLaVA-NeXT: Stronger LLMs Supercharge Multimodal Capabilities in the Wild},
    url={https://llava-vl.github.io/blog/2024-05-10-llava-next-stronger-llms/},
    author={Li, Bo and Zhang, Kaichen and Zhang, Hao and Guo, Dong and Zhang, Renrui and Li, Feng and Zhang, Yuanhan and Liu, Ziwei and Li, Chunyuan},
    _month={May},
    year={2024}
}

@inproceedings{radford2021learning,
  title={Learning transferable visual models from natural language supervision},
  author={Radford, Alec and Kim, Jong Wook and Hallacy, Chris and Ramesh, Aditya and Goh, Gabriel and Agarwal, Sandhini and Sastry, Girish and Askell, Amanda and Mishkin, Pamela and Clark, Jack and others},
  booktitle=ICML, 
  year={2021},
  _organization={PmLR}
}

@misc{chatgpt_citation,
  author = {OpenAI},
  title = {{ChatGPT}},
  howpublished = {\url{https://chat.openai.com/chat}},
  year = {2023},
}

@inproceedings{li2023blip,
  title={Blip-2: Bootstrapping language-image pre-training with frozen image encoders and large language models},
  author={Li, Junnan and Li, Dongxu and Savarese, Silvio and Hoi, Steven},
  booktitle=ICML,
  year={2023},
  organization={PMLR}
}

@article{wang2024qwen2,
  title={Qwen2-vl: Enhancing vision-language model's perception of the world at any resolution},
  author={Wang, Peng and Bai, Shuai and Tan, Sinan and Wang, Shijie and Fan, Zhihao and Bai, Jinze and Chen, Keqin and Liu, Xuejing and Wang, Jialin and Ge, Wenbin and others},
  journal={arXiv preprint arXiv:2409.12191},
  year={2024}
}

@inproceedings{chen2024sharegpt4v,
  title={Sharegpt4v: Improving large multi-modal models with better captions},
  author={Chen, Lin and Li, Jinsong and Dong, Xiaoyi and Zhang, Pan and He, Conghui and Wang, Jiaqi and Zhao, Feng and Lin, Dahua},
  booktitle=ECCV,
  pages={370--387},
  year={2024},
  organization={Springer}
}

@article{wada2025zina,
  title={ZINA: Multimodal Fine-grained Hallucination Detection and Editing},
  author={Wada, Yuiga and Matsuda, Kazuki and Sugiura, Komei and Neubig, Graham},
  journal={arXiv preprint arXiv:2506.13130},
  year={2025}
}

@inproceedings{gunjal2024detecting,
  title={Detecting and preventing hallucinations in large vision language models},
  author={Gunjal, Anisha and Yin, Jihan and Bas, Erhan},
  booktitle=AAAI,
  year={2024}
}

@article{wang2025internvl3,
  title={Internvl3. 5: Advancing open-source multimodal models in versatility, reasoning, and efficiency},
  author={Wang, Weiyun and Gao, Zhangwei and Gu, Lixin and Pu, Hengjun and Cui, Long and Wei, Xingguang and Liu, Zhaoyang and Jing, Linglin and Ye, Shenglong and Shao, Jie and others},
  journal={arXiv preprint arXiv:2508.18265},
  year={2025}
}

\appendix

\section{Appendix}
\label{sec:appendix}

\subsection{Additional Dataset Information}

\begin{figure}[h]
    \centering
   \includegraphics[width=0.5\textwidth]{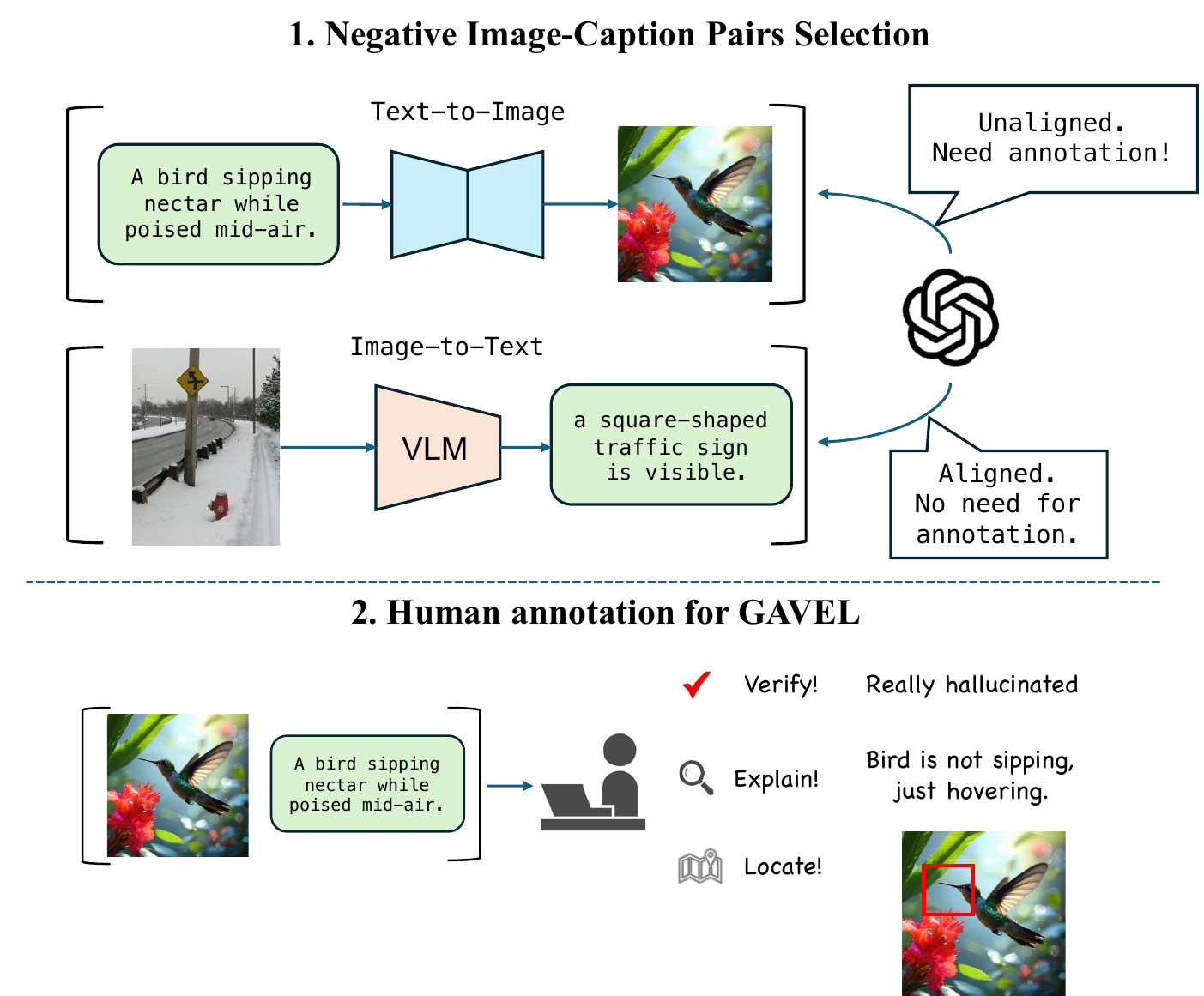}
    \caption{Overview of dataset construction pipeline.}
    \label{fig:data_con}
\end{figure}

\noindent
\noindent\textbf{Payment to Annotators.} The annotations were conducted by a professional annotation company. We compensated the annotators at a rate of approximately \$0.3 per annotated pair, based on the exchange rate at the time of the study.

\noindent\textbf{Dataset Construct Pipeline.}
Due to space limitations, the main paper describes the dataset construction process only in text. Here, we provide an additional illustration of the construction pipeline in Figure~\ref{fig:data_con}. The pipeline consists of two main stages: negative image--caption pair selection and human annotation.
In the first stage, we generate candidate image--caption pairs from two directions. For text-to-image generation, a caption is used to synthesize an image, and the resulting pair is evaluated for semantic alignment. For image-to-text generation, a VLM is prompted to generate a caption for a given image, and the generated pair is similarly assessed. We then use GPT-5 mini to filter these candidates according to their alignment scores. Pairs judged to be well aligned are discarded, while low-alignment pairs are regarded as likely hallucinated examples and forwarded to human annotators.
In the second stage, annotators verify whether each candidate pair is truly hallucinated. For incorrect pairs, they provide a textual explanation of the misalignment and localize the corresponding image region with a bounding box. This procedure produces the final annotations used in GAVEL.

\begin{figure}[t]
    \centering
   \includegraphics[width=0.48\textwidth]{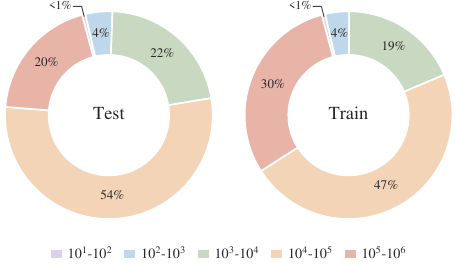}
    \caption{Distribution of bounding box areas in the training and test sets. The annular sectors indicate the proportions of bounding boxes in different area ranges measured in pixels$^2$.}
    \label{fig:bbx_size}
\end{figure}

\noindent
\textbf{Analysis of bounding boxes size.}
As shown in Fig.~\ref{fig:bbx_size}, bounding-box sizes are grouped into logarithmic intervals. Most instances are concentrated in the $10^4$--$10^5$ pixel range, accounting for 47\% of the training set and 54\% of the test set. In addition, bounding boxes in the $10^5$--$10^6$ pixel range account for 30\% and 20\% of the training and test sets, respectively, while extremely small bounding boxes below $10^3$ pixels constitute only a small fraction. Overall, the training and test sets exhibit similar bounding-box size distributions.

\subsection{Baseline Details}
We build our model upon the VisionLLM training framework and extend it with
task-specific coordinate tokens for spatial prediction. Instead of introducing
an additional bounding-box regression head, we formulate localization as an
autoregressive token generation problem.

Specifically, we augment the original tokenizer with a discrete coordinate
vocabulary
\begin{equation}
    \mathcal{P} = \{\tau_i \mid i \in [-512, 512]\},
\end{equation}
where each $\tau_i$ corresponds to a textual token
\texttt{\textless p$i$\textgreater}. This results in 1025 additional coordinate
tokens. After adding these tokens to the tokenizer, we resize the model's token
embedding matrix to match the new vocabulary size. During supervised
fine-tuning, both the input token embeddings and the language modeling head are
kept trainable, allowing the model to learn representations and output
probabilities for the newly introduced coordinate tokens.

For each ground-truth bounding box, the annotation is first converted from
$(x,y,w,h)$ to $(x_1,y_1,x_2,y_2)$. Each coordinate is then normalized with
respect to the original image resolution and quantized into the coordinate-token
range. Given a coordinate value $v$ along an axis of size $S$, we compute its
discrete index as
\begin{equation}
    q(v, S) =
    \mathrm{round}
    \left(
    512 \cdot
    \mathrm{clip}
    \left(
    \frac{v - S/2}{S/2}, -1, 1
    \right)
    \right).
\end{equation}
Here, $S$ denotes the image width for horizontal coordinates and the image
height for vertical coordinates. A bounding box is then serialized as
\begin{equation}
    \left(
    \tau_{q(x_1,W)},
    \tau_{q(y_1,H)},
    \tau_{q(x_2,W)},
    \tau_{q(y_2,H)}
    \right),
\end{equation}
where $W$ and $H$ are the original image width and height, respectively.

During data construction, the bounding-box placeholder \texttt{\textless BBX\textgreater}
in the assistant response is replaced by the corresponding serialized coordinate
tokens. The user prompt, visual tokens, and assistant response are formatted
following the Qwen/VisionLLM chat format. The supervised fine-tuning loss is
computed only over the assistant response tokens, while tokens belonging to the
system and user prompts are masked out. Thus, the model is trained to generate
both natural language responses and discrete spatial coordinates in a unified
autoregressive manner.

\subsection{LLM-as-a-Judge Evaluation Prompt}
\label{app:judge_prompt}

We use an LLM-as-a-judge protocol to evaluate the semantic consistency
between the predicted hallucination analysis and the ground-truth annotation.
The judge assigns a score from 1 to 5, where 5 indicates full semantic
consistency and 1 indicates an irrelevant or incorrect prediction. The judge
is required to output only a valid JSON object containing the score and a
short reason.

The system prompt is:

\begin{lstlisting}[style=promptstyle]
You are an evaluator. Output only a valid JSON object.
\end{lstlisting}

For samples with hallucination annotations, we use the following prompt:

\begin{lstlisting}[style=promptstyle]
Compare the following two texts:

Prediction:
{prediction}

Ground truth:
{ground_truth}

Consider whether the prediction agrees with the ground truth in meaning,
including the hallucination conclusion and the explanation.

Rate the prediction from 1 to 5:
5: fully consistent.
4: mostly consistent with only minor differences.
3: partially consistent but missing important details.
2: weakly related or largely inconsistent.
1: irrelevant or wrong.

Reply with a JSON object only:
{
  "score": 1-5,
  "reason": "one short sentence"
}
\end{lstlisting}

For samples whose ground-truth label is no hallucination, we use the
following prompt:

\begin{lstlisting}[style=promptstyle]
The ground truth hallucination label is:
NO HALLUCINATION.

This means the prediction should clearly indicate that there is no
hallucination and that the information is grounded or correct.

Prediction:
{prediction}

Judge only whether the prediction correctly expresses no hallucination
or an equivalent meaning, such as no error, description matches the image,
or everything is correct.

Ignore wording differences and focus on whether it clearly indicates
no hallucination.

Rate the prediction from 1 to 5:
5: clearly and correctly states no hallucination, or that the information
   is grounded and correct.
4: basically indicates no hallucination but with minor ambiguity.
2-3: unclear or ambiguous about hallucination status.
1: clearly wrong, e.g., claims hallucination or provides hallucinated
   or incorrect information.

Reply with a JSON object only:
{
  "score": 1-5,
  "reason": "one short sentence"
}
\end{lstlisting}

In implementation, both the prediction and ground-truth text are truncated to
at most 1500 characters before being provided to the judge.

\subsection{Computational Resources}

All training experiments were conducted on a computing cluster with
16 NVIDIA A100 GPUs, each equipped with 40 GB of GPU memory.
Inference was performed on a single NVIDIA A100 GPU with 80 GB of memory.

\subsection{Additional Qualitative Examples}
\label{app:qualitative_examples}

\begin{figure*}[h]
    \centering
   \includegraphics[width=0.98\textwidth]{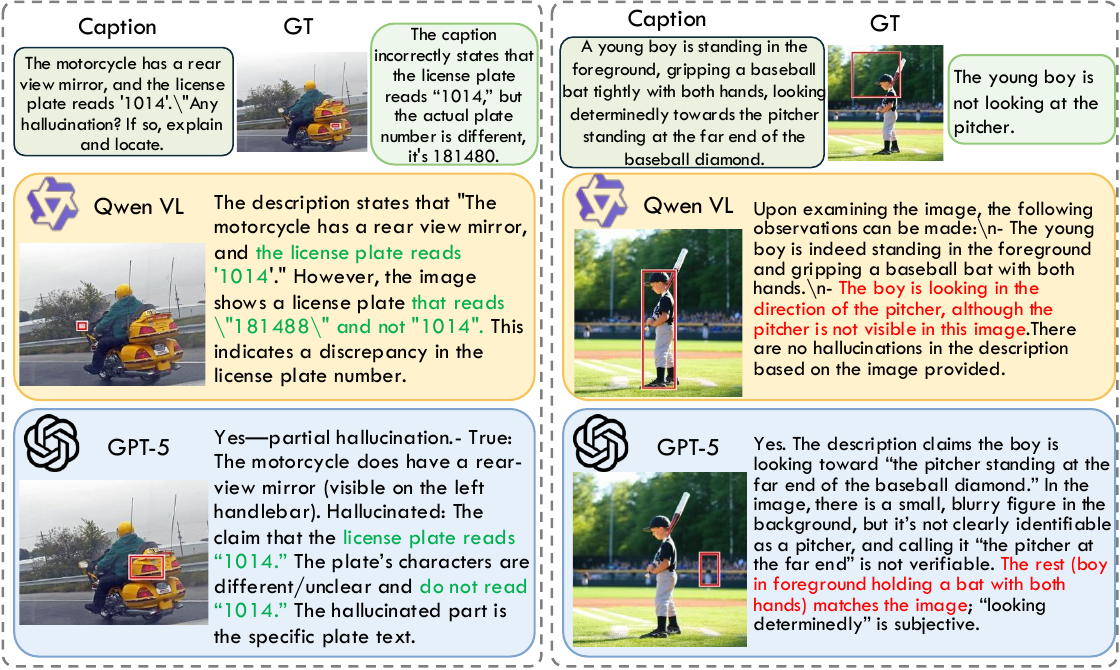}
    \caption{Additional qualitative examples from GAVEL comparing Qwen-VL and GPT-5 with ground-truth hallucination annotations. }
    \label{fig:vis_1}
\end{figure*}

\begin{figure*}[h]
    \centering
   \includegraphics[width=0.98\textwidth]{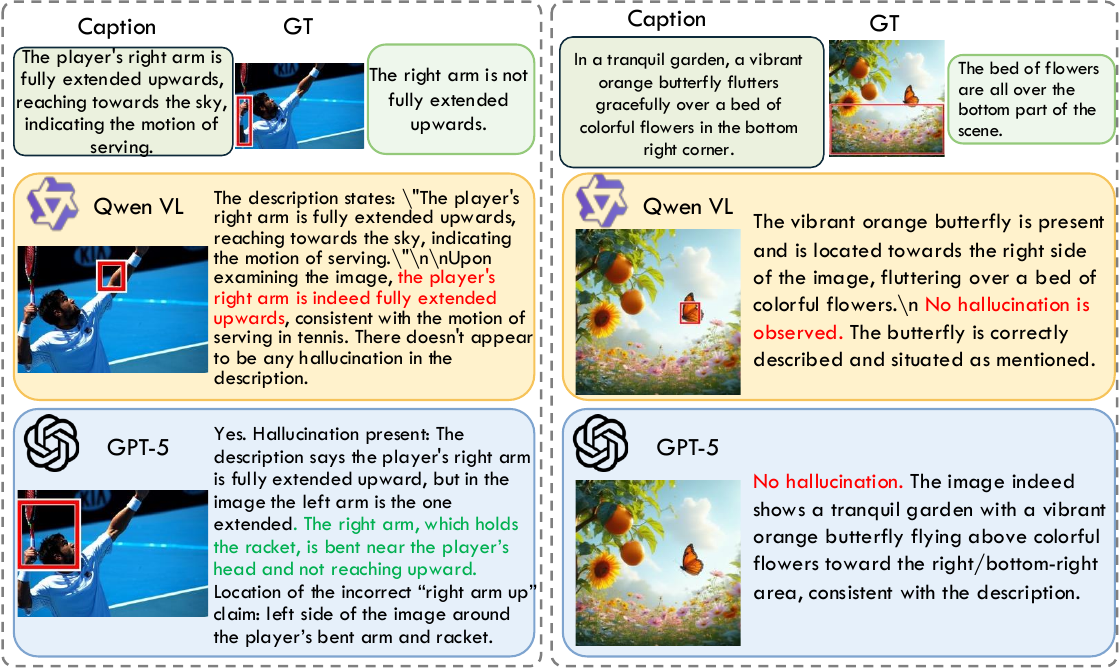}
    \caption{Additional qualitative examples from GAVEL comparing Qwen-VL and GPT-5 with ground-truth hallucination annotations. }
    \label{fig:vis_2}
\end{figure*}

\begin{figure*}[h]
    \centering
   \includegraphics[width=0.98\textwidth]{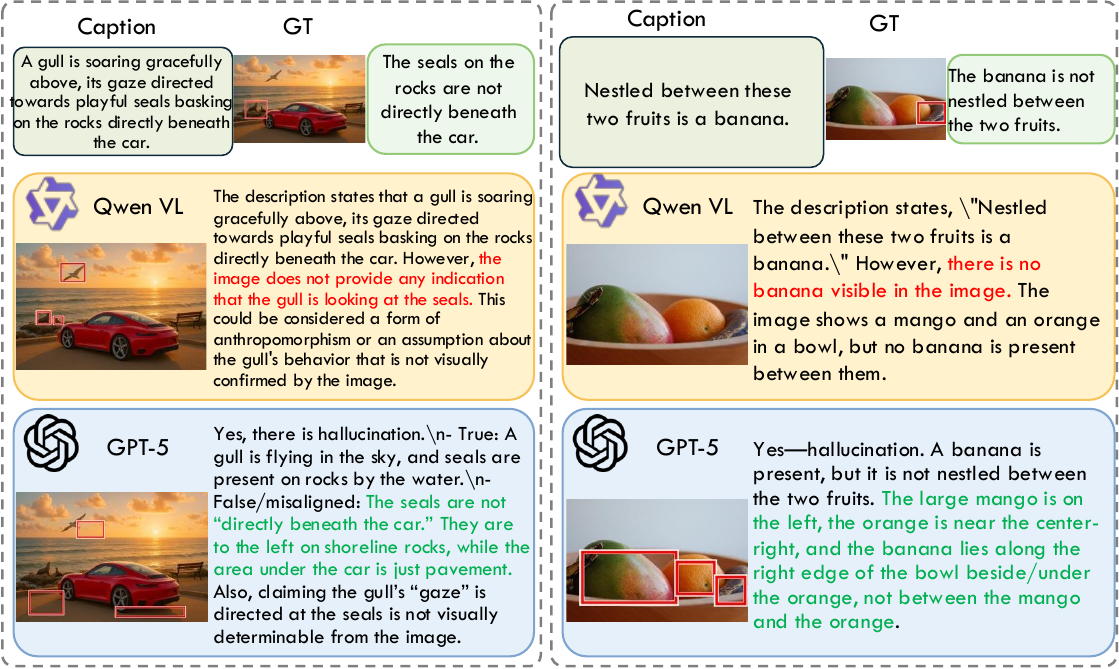}
    \caption{Additional qualitative examples from GAVEL comparing Qwen-VL and GPT-5 with ground-truth hallucination annotations. }
    \label{fig:vis_3}
\end{figure*}

\begin{figure*}[h]
    \centering
   \includegraphics[width=0.98\textwidth]{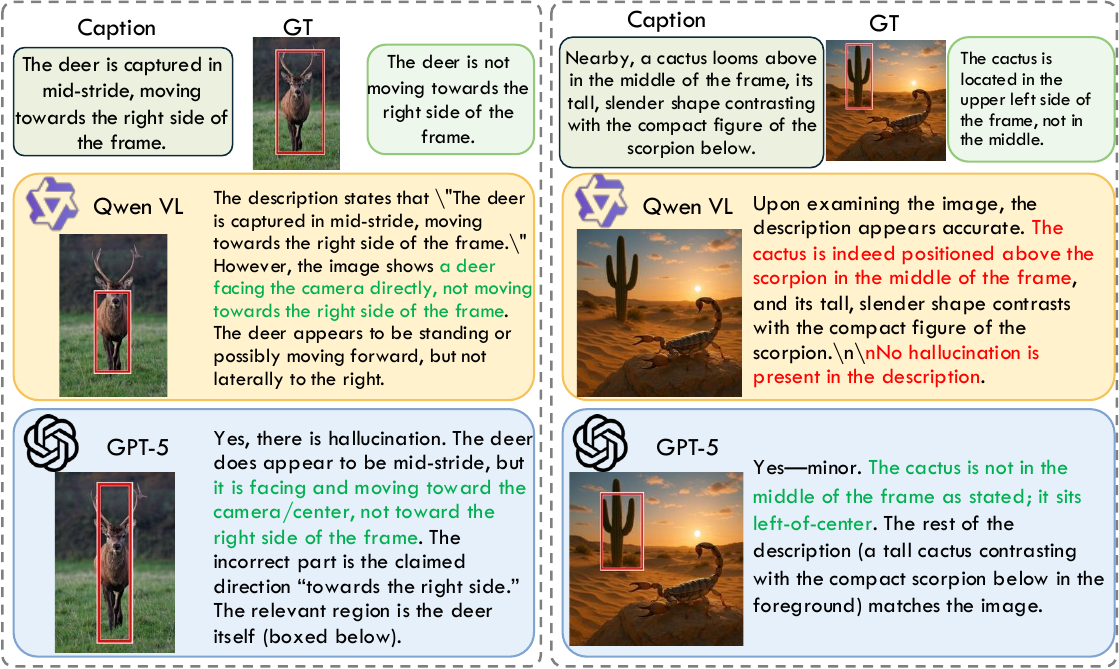}
    \caption{Additional qualitative examples from GAVEL comparing Qwen-VL and GPT-5 with ground-truth hallucination annotations. }
    \label{fig:vis_4}
\end{figure*}

We provide additional qualitative examples in Figs.~\ref{fig:vis_1}--\ref{fig:vis_4} to further illustrate the fine-grained hallucination detection and localization setting in GAVEL. 
Each example contains an image--caption pair, the ground-truth hallucination annotation, and model responses from representative vision-language models. 
The responses include both textual explanations and localized visual evidence, allowing us to compare whether models can correctly identify hallucinated content, explain the inconsistency, and ground the relevant image regions.

% These examples show that current models exhibit different behaviors when handling fine-grained hallucinations. 
% Some models can correctly detect the hallucinated content but fail to localize the precise visual evidence, while others produce plausible explanations that are not fully consistent with the image or the ground-truth annotation. 
% Such cases highlight the importance of evaluating hallucination understanding beyond image-level correctness, and demonstrate the diagnostic value of GAVEL for analyzing both explanation faithfulness and visual grounding accuracy.

\end{document}